\pdfoutput=1

\documentclass[11pt]{article}

\usepackage{ACL2023}

\usepackage{times}
\usepackage{latexsym}
\usepackage{diagbox}
\usepackage{graphicx} 
\usepackage{amsmath} 
\usepackage{multirow} 
\usepackage{arydshln} 

\usepackage[T1]{fontenc}

\usepackage[utf8]{inputenc}

\usepackage{microtype}

\usepackage{inconsolata}

\title{Prompt- and Trait Relation-aware Cross-prompt Essay Trait Scoring}

 \author{Heejin Do$^{\star}$, Yunsu Kim$^{\star \dagger}$, Gary Geunbae Lee$^{\star \dagger}$ \\
         $^{\star}$Graduate School of AI, POSTECH \\$^{\dagger}$Department of Computer Science and Engineering, POSTECH \\
         \texttt{\{heejindo, yunsu.kim, gblee\}@postech.ac.kr}}

\begin{document}
\maketitle
\begin{abstract}
Automated essay scoring (AES) aims to score essays written for a given prompt, which defines the writing topic. Most existing AES systems assume to grade essays of the same prompt as used in training and assign only a holistic score. However, such settings conflict with real-education situations; pre-graded essays for a particular prompt are lacking, and detailed trait scores of sub-rubrics are required. Thus, predicting various trait scores of unseen-prompt essays (called cross-prompt essay trait scoring) is a remaining challenge of AES. In this paper, we propose a robust model: prompt- and trait relation-aware cross-prompt essay trait scorer. We encode prompt-aware essay representation by essay-prompt attention and utilizing the topic-coherence feature extracted by the topic-modeling mechanism without access to labeled data; therefore, our model considers the prompt adherence of an essay, even in a cross-prompt setting. To facilitate multi-trait scoring, we design trait-similarity loss that encapsulates the correlations of traits. Experiments prove the efficacy of our model, showing state-of-the-art results for all prompts and traits. Significant improvements in low-resource-prompt and inferior traits further indicate our model's strength.
\end{abstract}


\section{Introduction}
Automated essay scoring (AES) aims to score essays written for a specific prompt, which defines the writing instructions and topic. As a subordinate or alternative to human scorers, it has the advantages of fairness and low costs. Thus far, most AES systems have been built on the assumptions of grading essays on the same prompt used for training and only assigning an overall score, achieving noticeable growth \citep{taghipour2016neural, dong2017attention, yang2020enhancing, wang2022use}.

\begin{figure}[t]
    \centering
    \includegraphics[width=7.7cm]{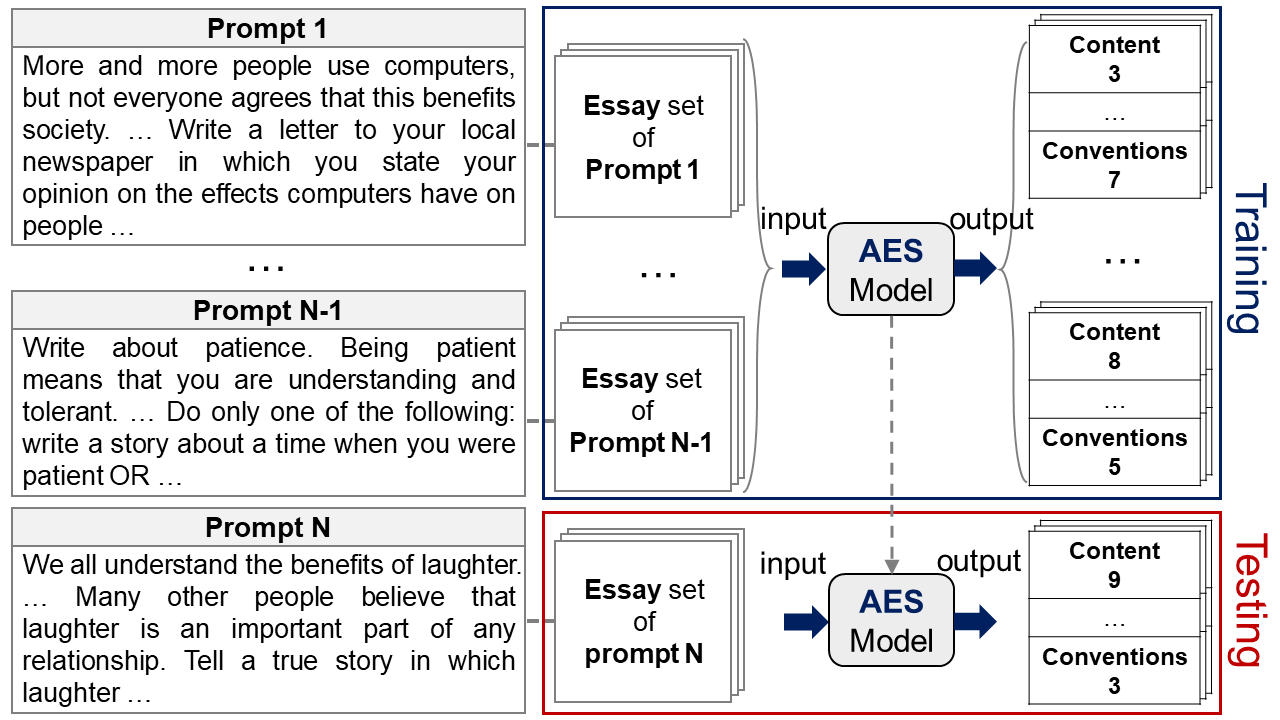}
    \caption{Cross-prompt essay trait scoring task}
    \label{fig1}
\end{figure}

However, such settings conflict with real-education systems, where pre-labeled essays for a specific prompt are not given, and in-depth feedback requires multiple trait scores. Acknowledging this, recent works have suggested cross-prompt models \citep{jin2018tdnn, li2020sednn, ridley2020prompt} that are tested using essays of unseen prompt, like zero-shot learning, and trait-scoring models \citep{mathias2020can, hussein2020trait, kumar2021many, he2022automated} that output multiple trait scores. Handling both settings (Figure~\ref{fig1}) is a direction for practical AES and yet has rarely been studied \citep{ridley2021automated}.

For a cross-prompt setting, using non-prompt-specific features that capture the general essay qualities such as length and readability is emphasized \citep{ridley2020prompt, uto2021review}. This is to avoid the model biased toward the prompts of trained essays, but the model fails to reflect any prompt-relevant information (e.g., whether the essay fits the prompt topic), inhibiting accurate scoring. For trait scoring, most methods extend holistic scoring models without particular consideration of trait properties. Both settings leave huge room for improvement.

In this paper, we propose a robust model, prompt- and trait relation-aware cross-prompt trait scorer (ProTACT), with the ultimate goal of improving AES for practical use. Attending to the prompt-relevant aspects and trait similarities leads to overcoming both cross-prompt and multi-trait settings.

To ensure that the model reflects prompt-relevant information, we introduce a novel architecture to obtain prompt-aware essay representation. Rather than only encoding the essay, we directly encode the prompt instruction and apply attention. This provides hints for scoring in cross-prompt settings since prompt content is always-given information, even for ungraded essays of new prompts. Furthermore, we suggest extracting the topic-coherence feature by applying the topic modeling mechanism latent Dirichlet allocation (LDA) \citep{blei2003latent}. This feature notifies essay coherence on a specific topic to the model without accessing labels.

To facilitate multi-trait scoring, we designed a trait-similarity loss that incorporates correlations between different trait scores. Practically, trait scores are not independent of one another; for example, both \textit{Prompt Adherence} and \textit{Content} traits evaluate prompt-relevant aspects of an essay. Finding strong correlations between trait scores, we mirror this for model training. Specifically, we penalize when the similarity of actual trait scores is over a threshold but that of predicted trait scores is low. This enhances the advantages of multi-trait learning by mutually assisting in different tasks.

We evaluate ProTACT with the widely used ASAP and ASAP++ datasets. ProTACT achieves state-of-the-art results, outperforming the baseline system \citep{ridley2021automated} for all QWK scores of traits and prompts. Significant improvements of 6.4\% on average and 10.3\% for the \textit{Content} trait are observed for a low-resource prompt, which performed poorly due to lacking similar-type training essays. This highlights the strength of ProTACT in the cross-prompt setting, overcoming the absence of pre-graded essays. Remarkably improved assessments for previously inferior traits further prove the effectiveness of multi-trait scoring. Codes and datasets are available on Github\footnote{\url{https://github.com/doheejin/ProTACT}}.

\section{Related Work}
AES studies mostly focus on the \textbf{prompt-specific holistic scoring} task. Aside from early machine learning-based regression or classification approaches \citep{landauer2003automated, attali2006automated, larkey1998automatic, rudner2002automated}, recent deep-learning-based methods for automatically learning essay representation are dominant. Notably, approaches that hierarchically represent essays from word- or sentence- to essay-level show competitive accuracy \citep{taghipour2016neural, dong2016automatic, dong2017attention}. Late attempts to fine-tune pre-trained models to develop more successful AES include \citet{yang2020enhancing}, who fine-tune BERT by combining regression and ranking loss, and \citet{wang2022use}, who suggest a multi-scale representation for BERT. \citet{zhang2019co} additionally encode source excerpts of source-dependent essays and suggest a co-attention. Our essay-prompt attention is distinct from theirs, as we encode the prompt rather than the source excerpt and apply attention differently.


Pointing out that previous successes in AES are far from real-world systems, few studies of the \textbf{cross-prompt} setting suggest methods of not examining target-prompt essays \citep{jin2018tdnn, li2020sednn, ridley2020prompt}. Considering the essay's semantic disparity by different prompts, the use of non-prompt-specific features of general essay qualities is highlighted in cross-prompt settings; \citet{ridley2020prompt} crafted the features of essay qualities, categorized as \textit{length-based}, \textit{readability}, \textit{text complexity}, \textit{text variation}, and \textit{sentiment}. However, they disregard the topic-coherence of the essay, which is an important consideration for grading \citep{miltsakaki2004evaluation}. To consider coherence during rating, we suggest a way of extracting the topic-coherence feature. 

To provide several trait scores that fit the sub-rubrics, a few \textbf{trait-scoring} studies have been proposed; however, they simply extend the existing holistic scoring methods by adding multi-output linear layers \citep{hussein2020trait} or using multiple trait-specific models \citep{mathias2020can, kumar2021many}. Emphasizing both the cross-prompt and trait scoring task, \citet{ridley2021automated} suggest a leading model for the \textbf{cross-prompt trait scoring} task. They extend the \citet{dong2017attention} model by setting multiple trait-specific layers and concatenating the features of \citet{ridley2021automated}. Despite achieving the best results on the task, the performance still lags far behind the prompt-specific holistic scoring. In addition, the performance gaps between traits and between target prompts are remarkable. We propose a novel architecture to improve cross-prompt trait scoring and thereby reduce the performance gap. 



\section{Model Description: ProTACT}

\begin{figure*}[htp]
    \centering
    \includegraphics[width=15cm]{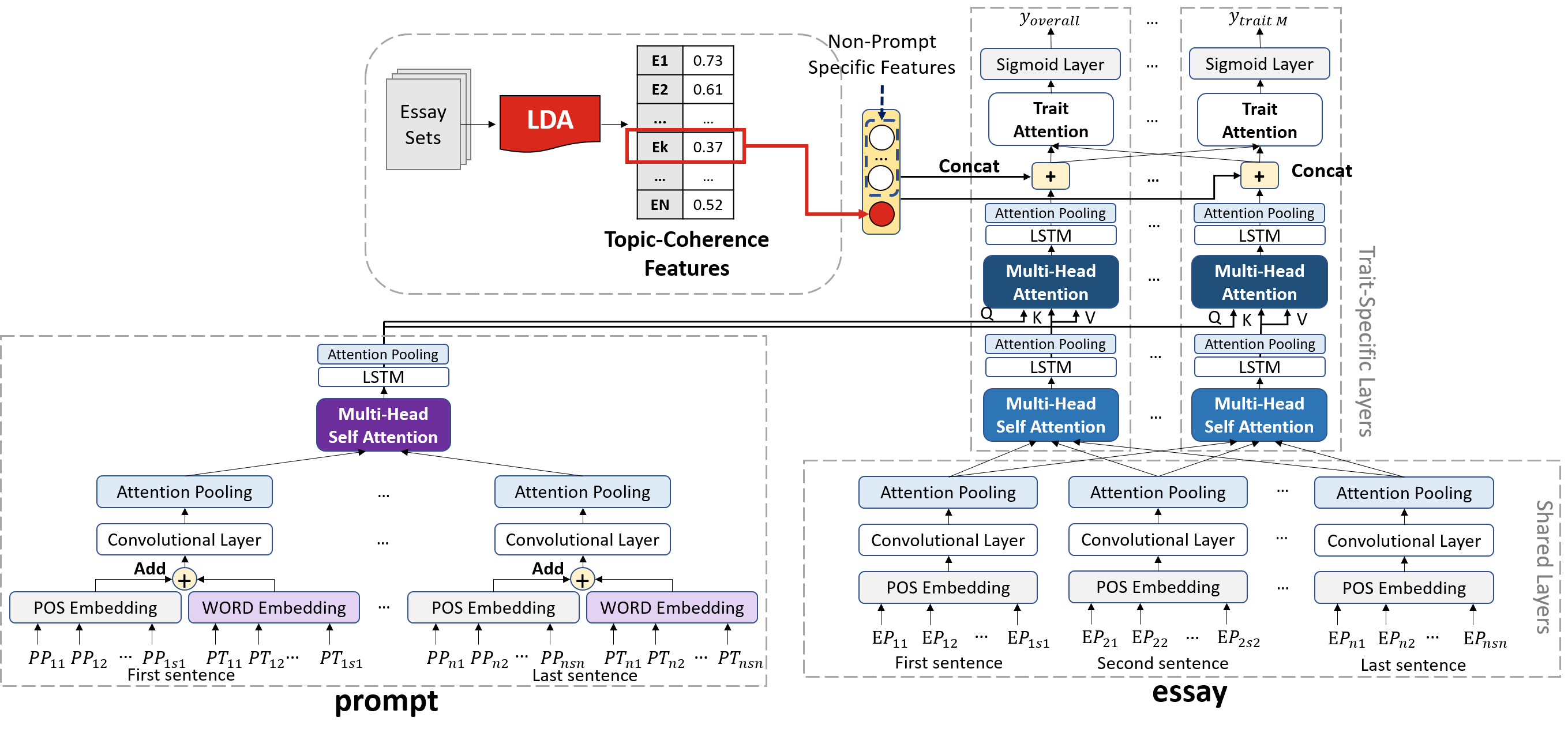}
    \caption{ProTACT model architecture.}
    \label{fig2}
\end{figure*}

To benefit from both automatically learning essay representations and precisely designed essay features, we combine both approaches. Therefore, ProTACT comprises two main parts: obtaining the prompt-aware essay representation and extracting the essay features (Figure~\ref{fig2}). The learned prompt-aware essay representation is concatenated with the pre-extracted essay features, constructing the final essay representation to score. The model is trained with the loss function that interpolates our trait-similarity loss and the mean squared error (MSE).

\subsection{Prompt-aware Essay Representation} 
We apply the hierarchical structure to encode the essay, first obtaining sentence-level representations and then a document-level representation. Hierarchically learning the document representation has proven effective for AES models, as it mirrors the essay structure that comprises sentences \citep{dong2017attention, ridley2020prompt, ridley2021automated}. 

To score multiple traits, we set separate trait-specific layers on top of the shared layers as the baseline model \citep{ridley2021automated}, but with different layer constructions. Shared layers and trait-specific layers are for sentence- and essay-level representations, respectively. To obtain {\itshape M} essay representations for {\itshape M} traits (including the overall score), {\itshape M} trait-specific modules exist. Sharing low-level layers enables information interchange between different traits, alleviating the data shortage caused by partial trait coverage.

To obtain prompt-aware essay representation for each trait, we introduce essay-prompt attention. Unlike existing methods that only encode the essay, we encode the prompt information in parallel and apply attention to the essay representation.

\paragraph{Essay Representation}

Instead of directly using word embedding, we use part-of-speech (POS) embedding for generalized representation, since doing so prevents overfitting to training data in cross-prompt settings \citep{jin2018tdnn, ridley2020prompt, ridley2021automated}. Each sentence is POS-tagged with the Python NLTK\footnote{\url{https://www.nltk.org/}} package, and the tagged words of each sentence are mapped to dense vectors. Then, to obtain \textbf{sentence-level representation}, the 1D convolutional layer followed by attention pooling \citep{dong2017attention} is applied for each sentence. The following equations explain the convolutional (Eq.~\ref{eq1}) and attention-pooling layers (Eqs.~\ref{eq2},~\ref{eq3},~\ref{eq4}):
\begin{eqnarray}
\mathbf{c}_{i} &=& f(\mathbf{W}_{c}\cdot[\mathbf{x}_{i}:\mathbf{x}_{i+h_{w}-1}]+\mathbf{b}_{c}) \label{eq1}\\
\mathbf{a}_{i} &=& \mathrm{tanh}(\mathbf{W}_{a}\cdot\mathbf{c}_{i}+\mathbf{b}_{a}) \label{eq2} \\
u_{i} &=& \frac{\mathrm{exp}(\mathbf{w}_{u}\cdot \mathbf{a}_{i})}{\sum \mathrm{exp}(\mathbf{w}_{u}\cdot \mathbf{a}_{j})} \label{eq3}\\
\mathbf{s} &=& \sum u_{i}\mathbf{c}_{i} \label{eq4}
\end{eqnarray}
where $\mathbf{c}_{i}$ is the feature representation after the convolutional layer, $\mathbf{W}_{c}$ is the weight matrix, $\mathbf{b}_{c}$ is the bias vector, and $\mathbf{h}_{w}$ is the window size of the convolutional layer. The final sentence representation $\mathbf{s}$ is obtained by the weighted sum where $\mathbf{u}_{i}$ is the attention weight, $\mathbf{a}_{i}$ is the attention vector, and $\mathbf{w}_{u}$ is the weight vector. $\mathbf{W}_{a}$ and $\mathbf{b}_{a}$ are the attention matrix and bias vector, respectively.

To examine each point of the long-range essays effectively, we first apply the multi-head self-attention \citep{vaswani2017attention} mechanism for the \textbf{essay-level representation}. Each trait-specific module takes the generated sentence-level representations as input and applies the multi-head self-attention. Consider the $j$-th trait score prediction task; the output of the previous layer, $S$, which is the matrix of sentence representations set as a query, key, and value:
\begin{eqnarray}
\mathrm{H_{i}^j}=\mathrm{Att}(SW_i^{j1}, SW_i^{j2}, SW_i^{j3}) \label{eq8}\\
\mathrm{MH}(S)^j=\mathrm{Concat}(\mathrm{H_{1}^j,..., H_{h}^j})W^{jO} \label{eq9}
\end{eqnarray}
where $\mathrm{Att}$ and $\mathrm{H_{i}}$ denote scaled-dot product attention and the $i$-th head, respectively, and $W_i^{j1}$, $W_i^{j2}$, and $W_i^{j3}$ are the parameter matrices. To the best of our knowledge, we are the first to apply the multi-head self-attention mechanism in both cross-prompt and trait-scoring settings. We hypothesize that this better models the structural aspect of the essay with the use of POS embedding and easily captures the relationship between different points of the essay from various perspectives.

Next, the recurrent layer of LSTM \citep{hochreiter1997long} is applied to the output:
\begin{eqnarray}
\mathbf{h}_t^j&=&\mathrm{LSTM}(m_{t-1}^j, m_{t}^j) \label{eq10}
\end{eqnarray}
where $j$ is the j-th trait score prediction task, $m^j$ is the concatenated output of the previous layer, and $\mathbf{h}_t^j$ denotes the hidden representation for the $j$-th task at time-step $t$. As LSTM captures sequential connections, directly applying it to a relation-encoded representation can lead to the sequential interpretation of relations \citep{li2018coherence}. This is followed by the attention pooling layer (Eqs.~\ref{eq2},~\ref{eq3},~\ref{eq4}).

\paragraph{Prompt Representation}
In practical education situations, grades are scored based on prompt instructions. Inspired by this, we encode the prompt instruction corresponding to each essay and make the model attend to it. Prompt representation is also obtained in the same order as the essay representation: embedding layer, convolutional layer with attention pooling, multi-head self-attention with LSTM, and attention pooling layer. However, to contain the contents of the prompt, we add the POS embedding with the pre-trained GloVe \citep{pennington2014glove} word embedding.

\paragraph{Essay-Prompt Attention} For the next step, essay-prompt attention is performed using a multi-head self-attention mechanism. The queries are set as the obtained prompt representation and the keys and values are set as the obtained essay representation. This allows every position of the essay to view sub-parts of the prompt; hence, essay-prompt attention captures the relationship between the essay and the prompt. Finally, the LSTM with attention pooling layer is applied to obtain the prompt-aware essay representation, $\mathbf{pa}^j$, for each $j$-th task.

\paragraph{Final Prediction} The essay representation is subsequently concatenated with pre-engineered features. As in the baseline model, we also use the non-prompt-specific features of PAES \citep{ridley2020prompt} that are exquisitely engineered to represent general essay quality in various aspects. However, we additionally concatenate our own feature of topic coherence. The feature vectors, $\mathbf{f}$, are then concatenated with each trait prediction, $\mathbf{pa}^j$: $\mathbf{con}^j = [\mathbf{pa}^j;\mathbf{f}]$.

Then, the trait-attention defined in \citet{ridley2021automated} is performed to attend to the representations of the other traits where $j=1, 2, \dots, M$:
\begin{eqnarray}
\mathbf{A}&=&[\mathbf{con}^1,\dots,\mathbf{con}^M] \label{eq10}\\
v_i^j&=&\frac{\mathrm{exp}(\mathrm{score}(\mathbf{con}^j, \mathbf{A}_{-j,i}))}{\sum_l \mathrm{exp}(\mathrm{score}(\mathbf{con}^j, \mathbf{A}_{-j,l}))}\label{eq12}\\
\mathbf{t}^j&=&\sum v_i^j \mathbf{A}_{-j,i} \label{eq13} \\
\mathbf{final}^j&=& [\mathbf{con}^j; \mathbf{t}^j] \label{eq13} 
\end{eqnarray}
where $\mathbf{A}$ is a concatenation of the representations for each trait prediction; $\mathbf{A}_{-j}$ indicates the masking of the target trait's representation; $v_i^j$ is the attention weight for the $i$-th trait; $\mathbf{t}^j$ is the attention vector. The final representation, $\mathbf{final}^j$, for each trait prediction is obtained by concatenating $\mathbf{con}^j$ and $\mathbf{t}^j$. Lastly, the final trait score, $\hat{y}^j$ is obtained by applying a linear layer with the sigmoid function $\hat{y}^j = \mathrm{sigmoid}(\mathbf{w}_y^j \cdot \mathbf{final}^j  + b_y^j)$. Here, $\mathbf{w}_y^j$ is a weights vector and $\mathbf{b}_y^j$ is a bias.

\begin{table}[t]
\centering
\scalebox{
0.73}{
\begin{tabular}{c|l|c}
\hline
Essay ID  & Topic Distribution [(Topic, Prob)] & \textbf{TC}\\
\hline
1 & [(0, \textbf{0.8337}), (5, 0.16295)] & 0.8337\\
2 & [(0, \textbf{0.7541}), (1, 0.0472), (5, 0.1472), ...] & 0.7541 \\
...  & ... & ... \\
11194  & [(2, 0.0477), (5, \textbf{0.8701}), (6, 0.0727)] & 0.8701\\
11195 & [(0, 0.0705), (2, 0.0664), (5, \textbf{0.8405}), ...] & 0.8405\\
\hline
\end{tabular}}
\caption{\label{tab1}
Example of the extracted features by LDA for each essay (\textbf{TC} denotes the Topic-coherence feature).}
\end{table}

\subsection{Topic-Coherence Feature}
\label{sec3.2}
To complement the existing non-prompt-specific features, in which prompt-related information is entirely excluded, we suggest using the LDA topic modeling mechanism. Looking at the document sets with the number of topics as a hyper-parameter, LDA identifies the topics and topic distributions for each document. Therefore, it can find out how an essay is focused on a particular topic, considering essays as documents. Since only essays are used without labels, features can be extracted even for new prompt essays in cross-prompt situations. 

Specifically, given the essay sets written for $N$ prompts, we apply LDA by setting the number of topics as $N$ to obtain the topic distribution for each essay (Table~\ref{tab1}); having multiple topics with low probabilities indicates lacked focus on a single topic, while the presence of a high-probability topic implies high focus on a certain topic. Then, the highest topic rate among the topic distributions for each essay is extracted as the topic-coherence feature. LDA is conducted separately for each training set since target-prompt essays should not be seen in training, eg., the training set of target-prompt 1 only includes prompts 2--8 essays. For testing, target-prompt essays are also used for extraction.

\begin{figure}[t]
    \centering
    \includegraphics[width=7.7cm]{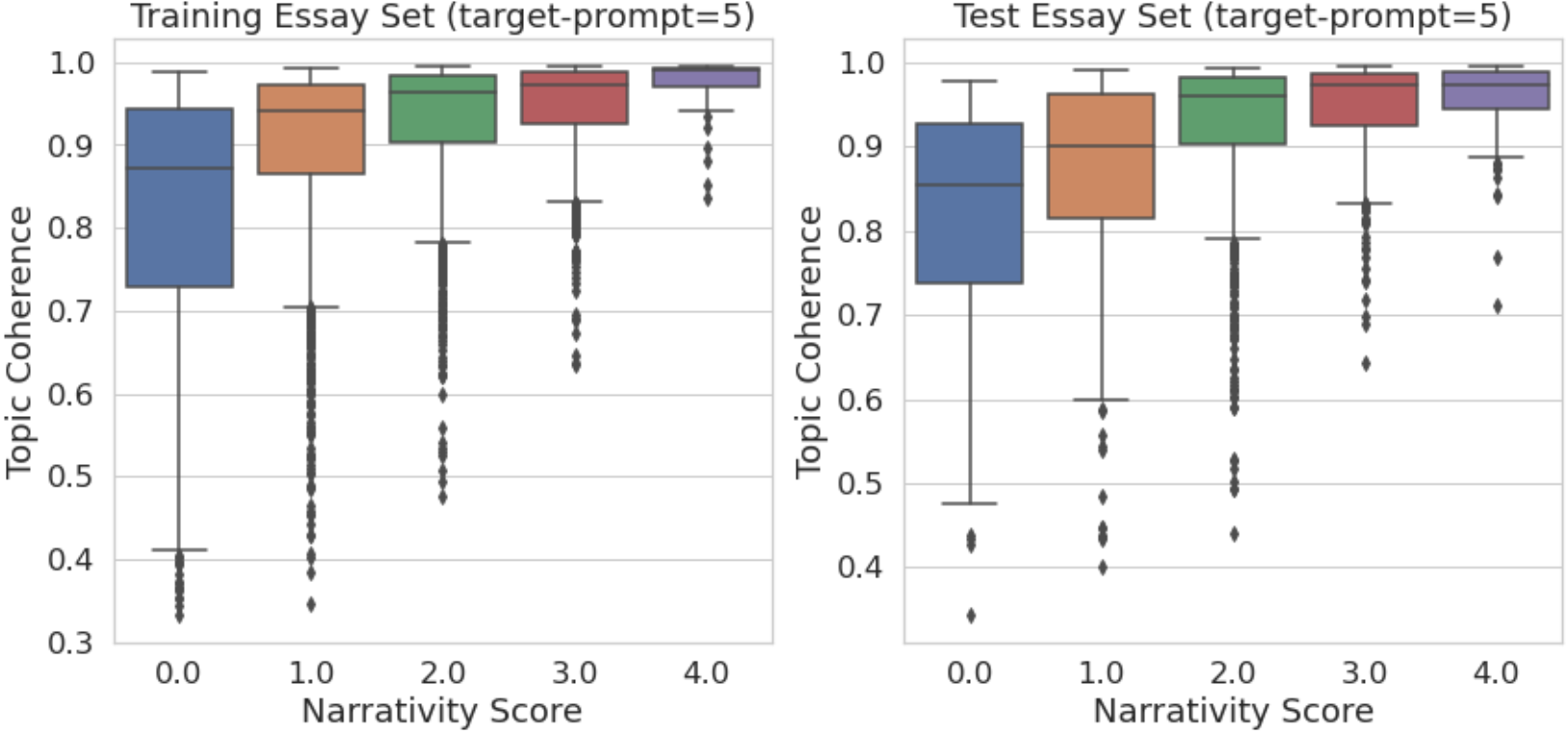}
    \caption{Box plot distributions of topic-coherence features of essays according to their \textit{Narrativity} trait scores.}
    \label{fig3}
\end{figure}

\paragraph{Does the Feature Imply Topic-coherence?} To examine whether our feature connotes the topic-coherence, we investigate the extracted feature's distribution with labeled \textit{Narrativity} trait score on our dataset, which is the attribute for evaluating the essay's coherence to the prompt, with integers 0--4. We plot the case when the target-prompt is 5, as it has the most essays among the prompts for which \textit{Narrativity} trait is evaluated. Figure~\ref{fig3} shows box plot\footnote{seaborn (\url{https://seaborn.pydata.org/}) is used.} distributions of extracted features according to essays' \textit{Narrativity} trait scores. The plotted training set only includes prompts 3,4, and 6 essays because only prompts 3--6 have labeled \textit{Narrativity} score. The greater distribution of high topic coherence at higher \textit{Narrativity} scores indicates that our feature reflects the essay's actual topic coherence. It is noteworthy that the test set shows similar trends in that our feature can give direct hints about consistency when scoring unseen prompt essays.



\paragraph{Does the Topic Correspond to Each Prompt?}
We further investigate the probability of the same prompt’s essays having the same highest topic, in each training set (Table~\ref{tab2}). Each \texttt{Set} denotes the training essay set of the target-prompt $n$, where LDA is separately applied. The left index of \texttt{Pr1--7} denotes different prompts by the \texttt{Set} since each target prompt is excluded. For example, index \texttt{Pr1} of \texttt{Set1} denotes the probability of prompt 2 essays having the same highest topic. Overall high probabilities imply that topics extracted by LDA are strongly related to the actual prompts, further notifying that our feature allows the model to recognize prompt relevance even in the cross-prompt setting.


\begin{table}[t]
\centering
\scalebox{
0.68}{
\begin{tabular}{c|c|c|c|c|c|c|c|c}
\hline
  & Set1 & Set2 & Set3 & Set4 & Set5 &  Set6 & Set7 & Set8\\
\hline
Pr1 & 0.996 & 1.000 & 0.685 & 0.680 & 1.000 & 0.979 & 0.885 & 0.996\\
Pr2 & 0.985 & 0.984 & 0.994 & 0.994 & 0.993 & 0.996 & 0.994 & 0.996 \\
Pr3  & 0.976 & 0.987 & 0.980 & 0.981 & 0.982 & 0.978 & 0.987 & 0.981 \\
Pr4 & 0.995 & 0.994 & 0.996 & 0.996 & 0.985 & 0.985 & 0.981 & 0.986 \\
Pr5 & 0.999 & 0.998 & 0.998 & 0.999 & 0.998 & 0.994 & 0.997 & 0.996 \\
Pr6 & 0.841 & 0.975 & 0.994 & 0.995 & 0.992 & 0.964 & 0.998 & 0.998 \\
Pr7 & 0.936 & 0.535 & 0.996 & 0.994 & 0.994 & 0.965 & 0.996 & 0.978 \\
\hline
Avg & 0.961 & 0.925 & 0.949 & 0.948 & 0.992 & 0.980 & 0.977 & 0.990\\
\hline
\end{tabular}}
\caption{\label{tab2}
The probabilities of the essays of the same prompt have the same highest topic in each training set.}
\end{table}

\begin{figure}[t]
    \centering
    \includegraphics[width=7.7cm]{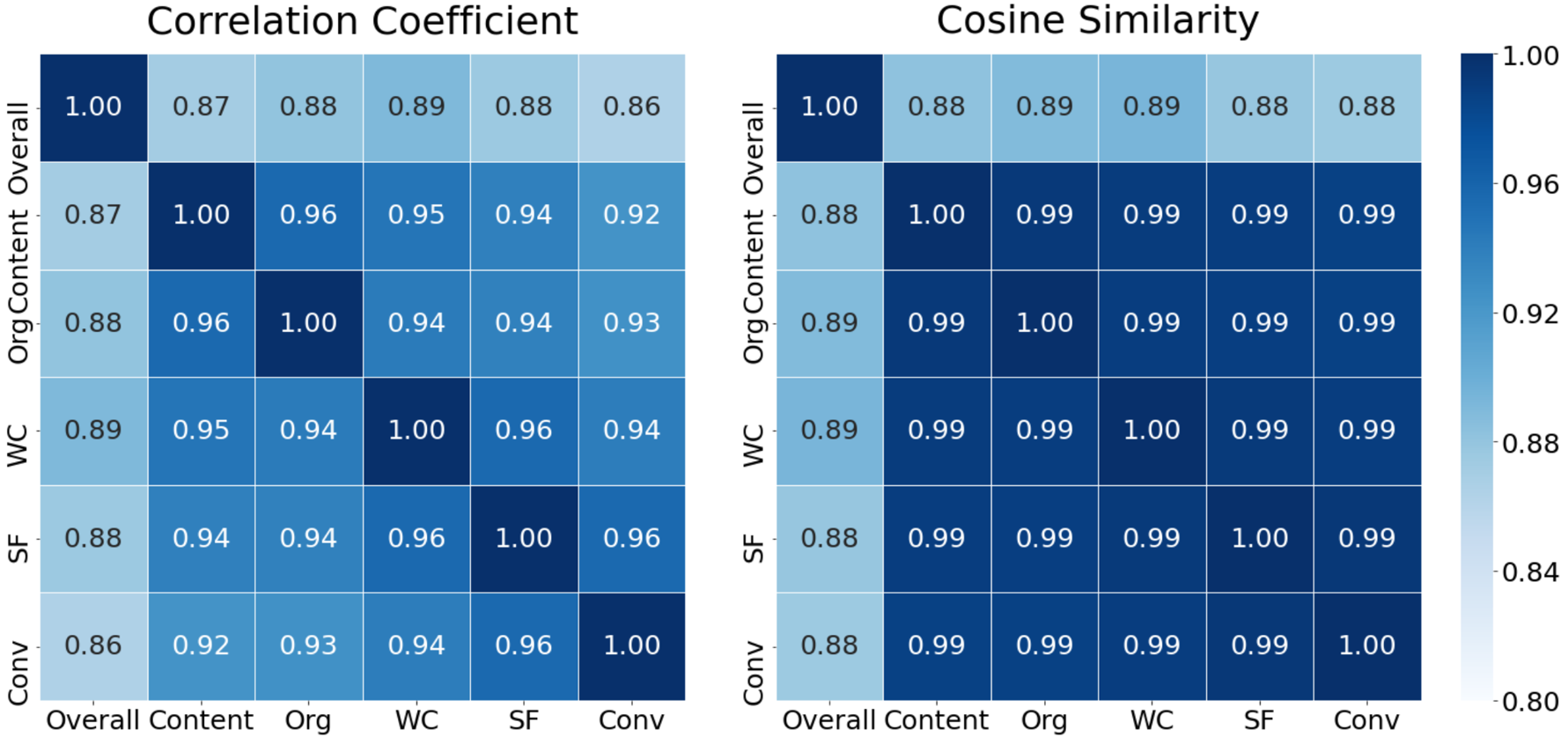}
    \caption{Correlation coefficients and cosine similarities between the ground-truth trait scores of prompt types 1, 2, and 8, which have the same trait composition.}
    \label{fig4}
\end{figure}

\begin{table*}
\centering
\scalebox{
0.69}{
\begin{tabular}{c|l|c|c|l}
\hline
Prompt & \multicolumn{1}{c|}{Essay Type} & Num of Essays & Avg Length  & \multicolumn{1}{c}{Traits}\\
\hline
1 & Argumentative & 1785 & 350  & Content, Word Choice, Organization, Sentence Fluency, Conventions \\
2 & Argumentative & 1800 & 350 & Content, Word Choice, Organization, Sentence Fluency, Conventions \\
3 & Source-Dependent & 1726 & 150  & Content, Prompt Adherence, Narrativity, Language \\
4 & Source-Dependent & 1772 & 150  & Content, Prompt Adherence, Narrativity, Language \\
5 & Source-Dependent & 1805 & 150  & Content, Prompt Adherence, Narrativity, Language \\
6 & Source-Dependent & 1800 & 150 & Content, Prompt Adherence, Narrativity, Language \\
7 & Narrative & 1569 & 300 & Content, Organization, Conventions \\
8 & Narrative & 723 & 650  & Content, Word Choice, Organization, Sentence Fluency, Conventions \\
\hline
\end{tabular}}
\caption{\label{tab3}
Summarization of the ASAP and ASAP++ combined dataset \cite{mathias2018asap++}.
}
\end{table*}

\subsection{Trait-Similarity Loss}
\label{sec3.3}
As in most AES systems, the existing cross-prompt trait scoring system is trained with the MSE loss. However, the only use of MSE loss disregards the correlations between different trait scores (Figure~\ref{fig4}). We integrate trait-relationship into the loss function, called the Trait-Similarity (TS) loss. In detail, when the similarity of the ground-truth trait score vectors is beyond the threshold, the model learning proceeds in the direction to increase the similarity of the predicted trait score vectors. The TS loss ($\mathrm{L_{ts}}$) is defined as follows:
\begin{eqnarray}
\mathrm{L_{ts}}({y},\hat{y})=\frac{1}{c}\sum_{j=2}^{M}\sum_{k=j+1}^{M}\mathrm{TS}(\hat{\mathbf{y}}_{j},\hat{\mathbf{y}}_{k},\mathbf{y}_{j},\mathbf{y}_{k}) \\
\mathrm{TS}=\begin{cases}
1-\mathrm{cos}(\hat{\mathbf{y}}_{j},\hat{\mathbf{y}}_{k}) &, \text{if } \mathrm{r}(\mathbf{y}_{j},\mathbf{y}_{k})\geq\delta\\
0 &, \text{otherwise} \end{cases}
\end{eqnarray}
where $\mathrm{cos}$ and $\mathrm{r}$ denote the cosine similarity and the Pearson correlation coefficient (PCC), respectively; $\delta$ is the threshold and $c$ is the number of calculated $TS$ that is not 0; $\mathbf{y}_j=[y_{1j},y_{1j},\cdots,y_{Nj}]$ is $j$-th ground-truth trait vector and $\hat{\mathbf{y}}_j=[\hat{y}_{1j},\hat{y}_{2j},\cdots,\hat{y}_{Nj}]$ is predicted trait vector. Note that \textit{Overall} trait ($j=1$) is excluded, as its score has relatively low correlations than other traits. The total loss, $\mathrm{L_{total}}$, is calculated as the interpolation of $\mathrm{L_{mse}}$ and $\mathrm{L_{ts}}$:
\[ \mathrm{L_{total}}(y,\hat{y})=\lambda\cdot\mathrm{L_{mse}}(y,\hat{y})+(1-\lambda)\cdot\mathrm{L_{ts}}(y,\hat{y}) \]
where the MSE loss is defined as, $
    \mathrm{L_{mse}}(y,\hat{y}) = \frac{1}{NM}\sum_{i=1}^{N}\sum_{j=1}^{M}(\hat{y}_{ij}-y_{ij})^2$, when predicting M trait scores for N essays and given the ground truth $y$ and prediction $\hat{y}$.
Note that TS loss of reflecting similarity between the traits is distinct from \citet{wang2022use}'s work of reflecting the similarity between the actual score and predicted score in a loss function for prompt-specific holistic scoring.

Given the entire trait set, $Y$, the specific trait set for each $i$-th training sample $Y^i$ differs depending on its prompt. Thus, for accurate calculation, masking to handle traits without gold scores is applied as $\mathbf{y}_{i}=\mathbf{y}_{i} \otimes mask_{i}$ and $\mathbf{\hat{y}}_{i}=\mathbf{\hat{y}}_{i} \otimes mask_{i}$. On the $i$-th essay, $mask_{ij}$ is computed for the $j$-th trait with the following function \cite{ridley2021automated}:
\begin{equation}
mask_{ij}=
\begin{cases}
1, & \text{if } Y_{j} \in Y^i \\
0, & \text{otherwise}
\end{cases}\end{equation}
\section{Experiment}
We experimented with the same dataset\footnote{\url{https://github.com/robert1ridley/cross-prompt-trait-scoring/tree/main/data}} as the baseline system, which is comprised of the publicly available Automated Student Assessment Prize (ASAP\footnote{\url{https://www.kaggle.com/c/asap-aes}}) and ASAP++\footnote{\url{https://lwsam.github.io/ASAP++/lrec2018.html}} datasets \citep{mathias2018asap++}. The original ASAP dataset contains eight prompts and their corresponding English-written essay sets, without personal information. Essays are assigned human-graded scores for their overall quality, and only essays of prompts 7 and 8 are assigned additional scores for several traits of scoring rubrics. Thus, the ASAP++ dataset, which has the same essay sets as ASAP but additionally graded trait scores for Prompts 1--6, is also utilized. Therefore, trait scores for prompts 1--6 are from the ASAP++, whereas trait scores for prompts 7 and 8 and overall scores for all prompts are from the ASAP dataset (Table~\ref{tab3}). For comparison, we exclude the \textit{Style} and \textit{Voice} attributes, which only appear in one prompt, as in the baseline model.

\begin{table*}[t]
\centering
\scalebox{
0.74}{
\begin{tabular}{l|cccccccc|c||c}
\hline
& \multicolumn{8}{c|}{\textbf{Prompts}} & & \\
\hline
\textbf{Model} & 1 & 2 & 3 & 4 & 5 & 6 & 7 & 8 & AVG & SD(↓) \\
\hline
PAES \citep{ridley2020prompt} & 0.605 & 0.522 & 0.575 & 0.606 & 0.634 & 0.545 & 0.356 & 0.447 & 0.536 & -\\
CTS \citep{ridley2021automated}& 0.623 & 0.540 & 0.592 & 0.623 & 0.613 & 0.548 & 0.384 & 0.504 & 0.553 & -\\
*CTS\small{-baseline} & 0.629 & 0.543 & 0.596 & 0.620 & 0.614 & 0.546 & 0.382 & 0.501 & 0.554 & 0.020 \\
\textbf{ProTACT} & \textbf{0.647} & \textbf{0.587} & \textbf{0.623} & \textbf{0.632} & \textbf{0.674} & \textbf{0.584} & \textbf{0.446} & \textbf{0.541} & \textbf{0.592} & 0.016 \\
\hline
\end{tabular}}
\caption{\label{tab4}
Average QWK scores over all traits for each \textbf{prompt}; \textit{SD} is the averaged standard deviation for five seeds, and \textbf{bold} text indicates the highest value.}
\smallskip
\smallskip
\smallskip
\scalebox{
0.74}{
\begin{tabular}{l|ccccccccc|c||c}
\hline
& \multicolumn{9}{c|}{\textbf{Traits}} & &\\
\hline
\textbf{Model} & Overall & Content & Org & WC & SF & Conv & PA & Lang & Nar & AVG & SD(↓)\\
\hline
PAES \citep{ridley2020prompt}& 0.657 & 0.539 & 0.414 & 0.531 & 0.536 & 0.357 & 0.570 & 0.531 & 0.605 & 0.527 & -\\
CTS \citep{ridley2021automated} & 0.67 & 0.555 & 0.458 & 0.557 & 0.545 & 0.412 & 0.565 & 0.536 & 0.608 & 0.545 & -\\
*CTS\small{-baseline} & 0.670 & 0.551 & 0.459 & 0.562 & 0.556 & 0.413 & 0.568 & 0.533 & 0.610 & 0.547 & 0.012\\
\textbf{ProTACT} & \textbf{0.674} & \textbf{0.596} & \textbf{0.518} & \textbf{0.599} & \textbf{0.585} & \textbf{0.450} & \textbf{0.619} & \textbf{0.596} & \textbf{0.639} & \textbf{0.586} & 0.009\\
\hline
\end{tabular}}
\caption{\label{tab5}
Average QWK scores over all prompts for each \textbf{trait} (WC: Word Choice; PA: Prompt Adherence; Nar: Narrativity; Org: Organization; SF: Sentence Fluency; Conv: Conventions; Lang: Language).}
\end{table*}

\paragraph{Validation and Evaluation}
For the overall training procedure, we applied the prompt-wise cross-validation that is used for the existing cross-prompt AES \citep{jin2018tdnn, ridley2020prompt, ridley2021automated}. In detail, essays of one prompt are set as test data while essays of other prompts are set as training data, which is repeated for each prompt. The development set of each case comprises essays of the same prompts as the training set. For the evaluation, we used Quadratic Weighted Kappa (QWK), the official metric for ASAP competition and most frequently used for AES tasks, which measures the agreement between the human rater and the system.

\paragraph{Training Details}
For a fair comparison, we maintained training details of the baseline model, other than those required by ProTACT. Out of the total 50 epochs, the one with the highest average QWK score for all traits in the development set was selected for the test. We set the dropout rate as 0.5, CNN filter and kernel size as 100 and 5, respectively, LSTM units as 100, POS embedding dimension as 50, and batch size as 10. We set two heads and the embedding dimension to 100 for multi-head attention. The total number of parameters is $2.76M$. For TS loss, $\delta$ of 0.7, and $\lambda$ of 0.7 are used. The RMSprop algorithm \citep{dauphin2015equilibrated} is used for optimization. The code is implemented in Tensorflow 2.0.0 and Python 3.7.11, and a Geforce RTX 2080Ti GPU card is used. Running the model five times with different seeds, $\{12, 22, 32, 42, 52\}$, the average scores represent the final scores. LDA is applied using the Gensim\footnote{\url{https://radimrehurek.com/gensim/}} library, specifying the number of prompts as the number of topics. Considering that each training and test uses an essay set of seven and eight prompts for LDA, the passes are set to 12 and 15, respectively.

\section{Results and Discussion}
\label{sec5}
The results clearly show that ProTACT outperforms the baseline CTS model for all prompts (Table~\ref{tab4}) and traits (Table~\ref{tab5}). In \citet{ridley2021automated}, PAES of the cross-prompt holistic scoring model is separately used for each trait scoring as a comparison of CTS, which is their proposed model. The {*CTS\small{-baseline}} is our implementation, with which we mainly compared our model for a fair comparison.

For target-prompt predictions (Table~\ref{tab4}), ProTACT achieved 3.8\% improvements on average. Compared to prompts 1 and 4, which already had high-quality predictions of 0.629 and 0.620, the other six prompts' predictions achieved larger improvement, reducing gaps between different prompts. This indicates that our methods provide more aid when predicting essays of a target prompt vulnerable to cross-prompt settings. 


We further investigated the low-resource prompt, which lacks similar-type essays in its training data (Table~\ref{tab6}). When predicting target-prompt 7, only 723 essays are of the same \textit{Narrative} type (prompt 8) in the training set (Table~\ref{tab3}). We compare their results with prompts 1,2, and 8, which contain all traits of prompt 7. ProTACT for target-prompt 7 achieved an average 6.4\% increment, and especially a 10.3\% increment for the \textit{Content} trait; the improvement rate is almost twice that of prompts 1, 2, and 8. This is noticeable given the severely inferior baseline target-prompt 7 predictions of all three trait scores, except \textit{Overall}. Another point to note is that prompts 1, 2, 8, and 7 all deal with long essays (Table~\ref{tab3}) that require strong encoding ability \citep{wang2022use}, but improved 4.2\% on average, implying the efficacy of our encoding strategy.

\begin{table}[t]
\centering
\scalebox{
0.69}{
\begin{tabular}{c|c|c|c|c|c|c}
\hline
Target & \textbf{Model}  & Overall & Content & Org & Conv & \textbf{Avg} \\
\hline
\multirow{3}{*}{\begin{tabular}{c} 1,2,8 \\ (avg) \end{tabular}} & 
*CTS\small{-baseline} & \textbf{0.679} & 0.523 & 0.535 & 0.490 & 0.557\\
& ProTACT & 0.673  & \textbf{0.585}  & \textbf{0.585}  & \textbf{0.523} & \textbf{0.592}\\
\cdashline{2-7}
& $\Delta$ & -0.6\% & 6.2\% & 5.0\% & 3.3\% & 3.5\% \\
\hline
\multirow{3}{*}{7} & 
*CTS\small{-baseline} & 0.720 & 0.398 & 0.231 & 0.179 & 0.382\\
& \textbf{ProTACT} & \textbf{0.735} & \textbf{0.501} & \textbf{0.315} & \textbf{0.232} & \textbf{0.446} \\
\cdashline{2-7}
& $\Delta$ & 1.5\% & 10.3\% & 8.4\% & 5.3\% & 6.4\% \\
\hline
\end{tabular}}
\caption{\label{tab6}
Comparison of QWK scores for Prompt 7 and Prompt 1,2,8 (averaged). \textit{Target} means target-prompt.}
\end{table}

For each trait scoring task (Table~\ref{tab5}), ProTACT achieved an average 3.9\% enhancement over the baseline system. In particular, noticeable improvements are shown in all traits except the \textit{Overall}, which already had considerably higher performance than other traits. Multiple trait-scoring tasks share information between layers, so inferior tasks might benefit more than superior tasks. Thus, ProTACT alleviates the data shortages in specific trait-scoring tasks caused by partial-trait coverage. 

\begin{table*}[t]
\centering
\scalebox{
0.74}{
\begin{tabular}{l|cccccccc|c||c}
\hline
& \multicolumn{8}{c|}{\textbf{Prompts}} & & \\
\hline
\textbf{Model} & 1 & 2 & 3 & 4 & 5 & 6 & 7 & 8 & AVG & SD(↓) \\
\hline
*CTS\small{-baseline} & 0.629 & 0.543 & 0.596 & 0.620 & 0.614 & 0.546 & 0.382 & 0.501 & 0.554 & 0.020\\
\hline
MSA & 0.635 & 0.561 & 0.594 & 0.617 & 0.617 & 0.557 & 0.404 & 0.533 & 0.565 & 0.017 \\
+ Essay-Prompt Att & 0.638 & 0.559 & 0.595 & 0.624 & 0.615 & 0.567 & 0.397 & 0.531 & 0.566 & 0.017 \\
+ TC feature & 0.639 & 0.581 & 0.618 & \textbf{0.634} & 0.657 & 0.580 & 0.436 & 0.525 & 0.584 & 0.015 \\
+ TS loss \small{(ProTACT)} & \textbf{0.647} & \textbf{0.587} & \textbf{0.623} & 0.632 & \textbf{0.674} & \textbf{0.584} & \textbf{0.446} & \textbf{0.541} & \textbf{0.592} & 0.016 \\
\hline
\end{tabular}}
\caption{\label{tab7}
Results of ablation studies. The average QWK scores over all traits for each \textbf{prompt}. \textit{MSA} denotes multi-head self-attention, \textit{TC feature} denotes the Topic-coherence feature.}
\smallskip
\smallskip
\smallskip
\scalebox{
0.74}{
\begin{tabular}{l|ccccccccc|c||c}
\hline
& \multicolumn{9}{c|}{\textbf{Traits}} & &\\
\hline
\textbf{Model} & Overall & Content & Org & WC & SF & Conv & PA & Lang & Nar & AVG & SD(↓)\\
\hline
*CTS\small{-baseline} & 0.670 & 0.551 & 0.459 & 0.562 & 0.556 & 0.413 & 0.568 & 0.533 & 0.610 & 0.547 & 0.012\\
\hline
MSA & 0.671 & 0.562 & 0.486 & 0.580 & 0.573 & 0.441 & 0.568 & 0.545 & 0.610 & 0.560 & 0.012 \\
+ Essay-Prompt Att & 0.671 & 0.565 & 0.477 & 0.582 & 0.574 & 0.435 & 0.573 & 0.550 & 0.618 & 0.561 & 0.012 \\
+ TC feature & 0.673 & 0.592 & 0.500 & 0.591 & 0.577 & 0.444 & 0.612 & 0.570 & 0.633 & 0.577 & 0.012 \\
+ TS loss \small{(ProTACT)} & \textbf{0.674} & \textbf{0.596} & \textbf{0.518} & \textbf{0.599} & \textbf{0.585} & \textbf{0.450} & \textbf{0.619} & \textbf{0.596} & \textbf{0.639} & \textbf{0.586} & 0.009\\
\hline
\end{tabular}}
\caption{\label{tab8}
Results of ablation studies. The average QWK scores over all prompts for each \textbf{trait}.}
\end{table*}

\begin{figure}[t]
    \centering
    \includegraphics[width=7.7cm]{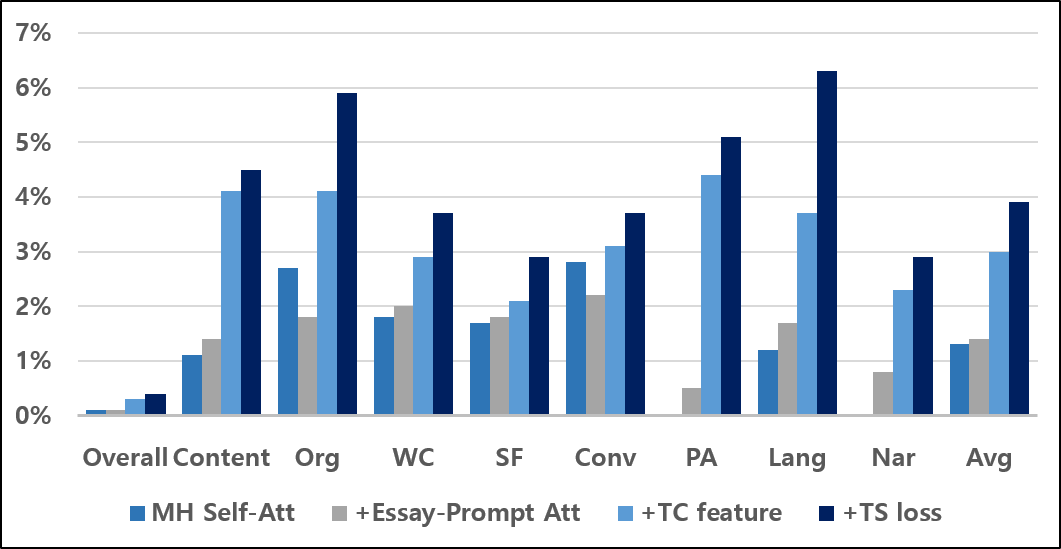}
    \caption{Improvement over the baseline model when incrementally applying each method of ProTACT.}
    \label{fig5}
\end{figure}

\subsection{Ablation Studies}
\label{ablation}
\paragraph{Incremental Analysis}
To explore the impact of each model component, we conducted an incremental analysis. Starting from encoding essay representation with a multi-head self-attention mechanism and using general essay features, we gradually added essay-prompt attention, topic-coherence feature, and TS loss. The results show both prompt- and trait-wise incremental advances (Tables~\ref{tab7},\ref{tab8}).

In particular, remarkable improvements on all prompts after applying \textit{+TC feature} prove that informing prompt-related knowledge facilitates scoring in cross-prompt settings (Table~\ref{tab7}). Figure~\ref{fig5} shows increases in trait scoring tasks over the baseline (Table~\ref{tab8}). The simple use of self-attention improves overall trait-scoring tasks, especially for syntactic traits such as \textit{Conventions} and \textit{Organization}, which evaluate overall grammatical writing conventions and essay structure, respectively. This matches our goal of multi-head self-attention capturing the structural and syntactic aspects. In contrast, supplementary use of essay-prompt attention somewhat decreases the scoring quality for those syntactic traits, yet particularly increases prompt-relevant traits such as \textit{Prompt-Adherence} and \textit{Narrativity}. Using the topic-coherence feature remarkably enhances scoring for \textit{Prompt Adherence} and \textit{Content} traits, which evaluates the essay's adherence to the topic and quantity of prompt-relevant text in the essay, respectively \citep{mathias2018asap++}. The results on typical coherence-related traits \citep{shin2022evaluating} prove that our feature explicitly supports related-aspect scoring and grows interpretability. Lastly, TS loss enhances all trait-scoring tasks, which shows the reflection of trait correlations boosts multi-trait joint learning.

\begin{table}[t]
\centering
\scalebox{
0.71}{
\begin{tabular}{l|c|c|c|c|c}
\hline
\diagbox[width=8em]{ }{$\delta$}& 0.5 & 0.6 & 0.7 & 0.8 & 0.9 \\
\hline
PCC & 0.583 & 0.585 & \textbf{0.586} & \textbf{0.586} & \textbf{0.586}  \\
Cosine Similarity & 0.585 & \textbf{0.586} & 0.584 & 0.585 & 0.585 \\
\hline
\end{tabular}}
\caption{\label{tab9}
Results of the TS loss with variations}
\end{table}

\paragraph{TS Loss with Variations}
To further optimize the TS loss, we have changed the criterion for the loss from PCC to cosine similarity. In addition, we experimented with the different values of the hyper-parameter $\delta$ for both conditions. Different $\delta$ values greater than 0.6 and condition change have little influence (Table~\ref{tab9}). Since the average correlation between trait scores is 0.87 and the cosine similarity is 0.97, no significant variation appeared when constraining the similarity over high values. 

\section{Conclusion}
We proposed a prompt- and trait relation-aware cross-prompt essay trait scorer (ProTACT) to improve AES in practical settings. Experimental results prove that informing prompt-relevant knowledge to the model assists the scoring of unseen prompt essays, and capturing trait similarities facilitates joint learning of multiple traits. Significant improvements in low-resource-prompt and inferior traits indicate the capacity to overcome the lacked pre-rated essays and strength in multi-trait scoring.

\section*{Limitations}
The limitations of our work can be summarized in three points. First, as mentioned in Section~\ref{sec5}, although a direct consideration of prompt information is helpful for related trait-scoring tasks, it may not be for irrelevant traits. Therefore, selectively applying each method depending on which traits are to score might further improve the model. Second, although the use of pre-engineered features, such as our topic-coherence feature, has the advantage of interpretability \citep{uto2020neural}, it requires additional engineering steps, as in other AES studies using hand-crafted features \citep{amorim2018automated, dascalu2017readerbench, nguyen2018argument, ridley2021automated}. Finally, despite the large improvements observed on the specific datasets ASAP and ASASP++, the model has not experimented on other datasets. Feedback Prize dataset\footnote{\url{https://www.kaggle.com/competitions/feedback-prize-2021/data}} is well-designed for scoring English-written argumentative writings with multiple trait labels, but the prompts are not defined; thus, it does not fit for cross-prompt AES. Essay-BR dataset \citep{marinho2022essay} contains essays on multiple prompts with labeled multiple trait scores. Thus, in future work, our proposed methods can be extended to multilingual cases of AES using the dataset. 

\section*{Ethics Statement}
We adhere to the ACL Code of Ethics. This work did not use any private datasets and did not contain any personal confidential information.

\section*{Acknowledgements}
This work was partly supported by Institute of Information \& communications Technology Planning \& Evaluation (IITP) grant funded by the Korea government (MSIT) (No.2019-0-01906, Artificial Intelligence Graduate School Program (POSTECH)) and MSIT (Ministry of Science and ICT), Korea, under the ITRC (Information Technology Research Center) support program (IITP-2023-2020-0-01789) supervised by the IITP (Institute for Information \& Communications Technology Planning \& Evaluation).

\bibliography{main}
\bibliographystyle{acl_natbib}

\appendix

\section{Detailed Ablation Studies}
\label{sub:appendix1}
In our main paper, we have conducted the incremental analysis in Section~\ref{ablation} to examine the effect of gradually adding each model component. The results have shown that adding the TC feature to the model, where multi-head self-attention and essay-prompt attention are applied, yields the greatest performance improvement. To closely investigate the individual contribution of the seemingly effectual TC feature, we now compare the results of separately adding the TC feature and essay-prompt attention (Table~\ref{tab11}). 

The noticeable point is that despite little overall improvements when separately applying essay-prompt attention and the TC feature, their simultaneous application leads to significantly increased performance. These results indicate our proposed methods can yield synergies when jointly applied.

\begin{table*}[t]
\centering
\scalebox{
0.73}{
\begin{tabular}{l|ccccccccc|c||c}
\hline
& \multicolumn{9}{c|}{\textbf{Traits}} & &\\
\hline
\textbf{Model} & Overall & Content & Org & WC & SF & Conv & PA & Lang & Nar & AVG & SD(↓)\\
\hline
MSA & 0.671 & 0.562 & 0.486 & 0.580 & 0.573 & 0.441 & 0.568 & 0.545 & 0.610 & 0.560 & 0.012 \\
MSA + Essay-Prompt Att & 0.671 & 0.565 & 0.477 & 0.582 & 0.574 & 0.435 & 0.573 & 0.550 & 0.618 & 0.561 & 0.012 \\
MSA + TC feature & 0.672 & 0.562 & 0.485 & 0.585 & 0.565 & 0.428 & 0.609 & 0.568 & 0.629 & 0.567 & 0.011 \\
MSA + Essay-Prompt Att + TC feature & \textbf{0.673} & \textbf{0.592} & \textbf{0.500} & \textbf{0.591} & \textbf{0.577} & \textbf{0.444} & \textbf{0.612} & \textbf{0.570} & \textbf{0.633} & \textbf{0.577} & 0.012 \\
\hline
\end{tabular}}
\caption{\label{tab11}
Results of detailed ablation studies. The average QWK scores over all prompts for each \textbf{trait}.}
\end{table*}

\begin{figure*}[t]
    \centering
    \includegraphics[width=13cm]{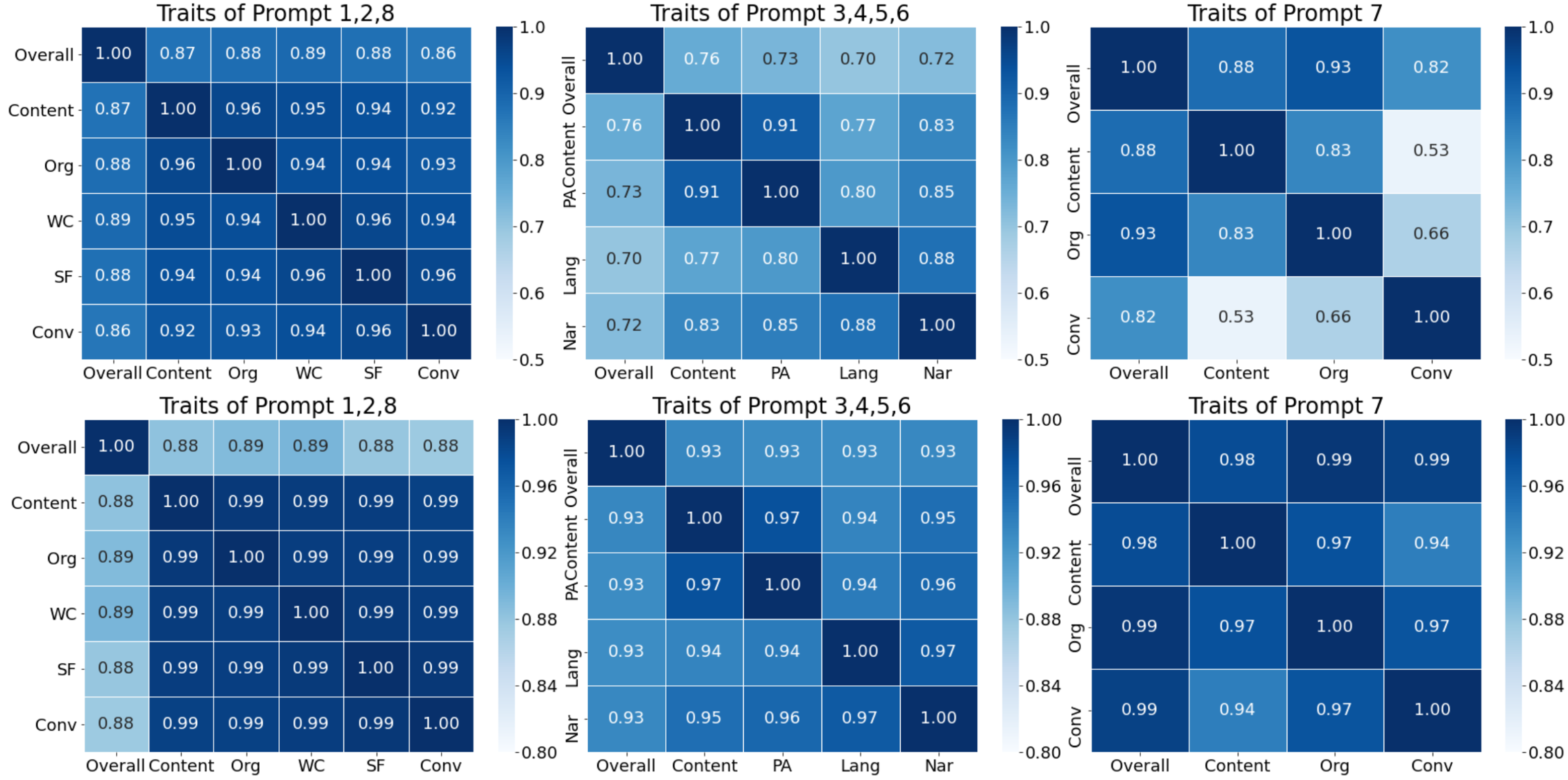}
    \caption{Correlation coefficient (1st row) and cosine similarity (2nd row) between ground-truth trait scores of all prompt types.}
    \label{fig6}
\end{figure*}

\section{Analysis of Trait Relationship}
\label{sub:appendix3}

In Section~\ref{sec3.3}, we showed the correlation coefficients and cosine similarities between the ground-truth trait scores of prompt types 1,2, and 8, which have the same trait composition. To further analyze relations between all different traits, we additionally examined trait scores of other prompts (Figure~\ref{fig6}). Likewise, we investigated the relationship between trait scores within prompts that are evaluated of the same traits. The correlation and cosine similarity results within the same prompt sets show similar tendencies, although the specific values are different. This explains the construction of our TS loss, which has criteria of correlation between actual trait scores while reflecting cosine similarities of predicted trait scores. Moreover, we find out higher similarities between prompt-related traits such as \textit{Prompt Adherence} and \textit{Content}. However, a relatively low association is observed for traits with distinctive evaluation rubrics, such as \textit{Conventions} and \textit{Content} traits. 

\section{Topic-coherence Feature and Related Traits}
\label{sub:appendix4}
\begin{figure}[t]
    \centering
    \includegraphics[width=7.8cm]{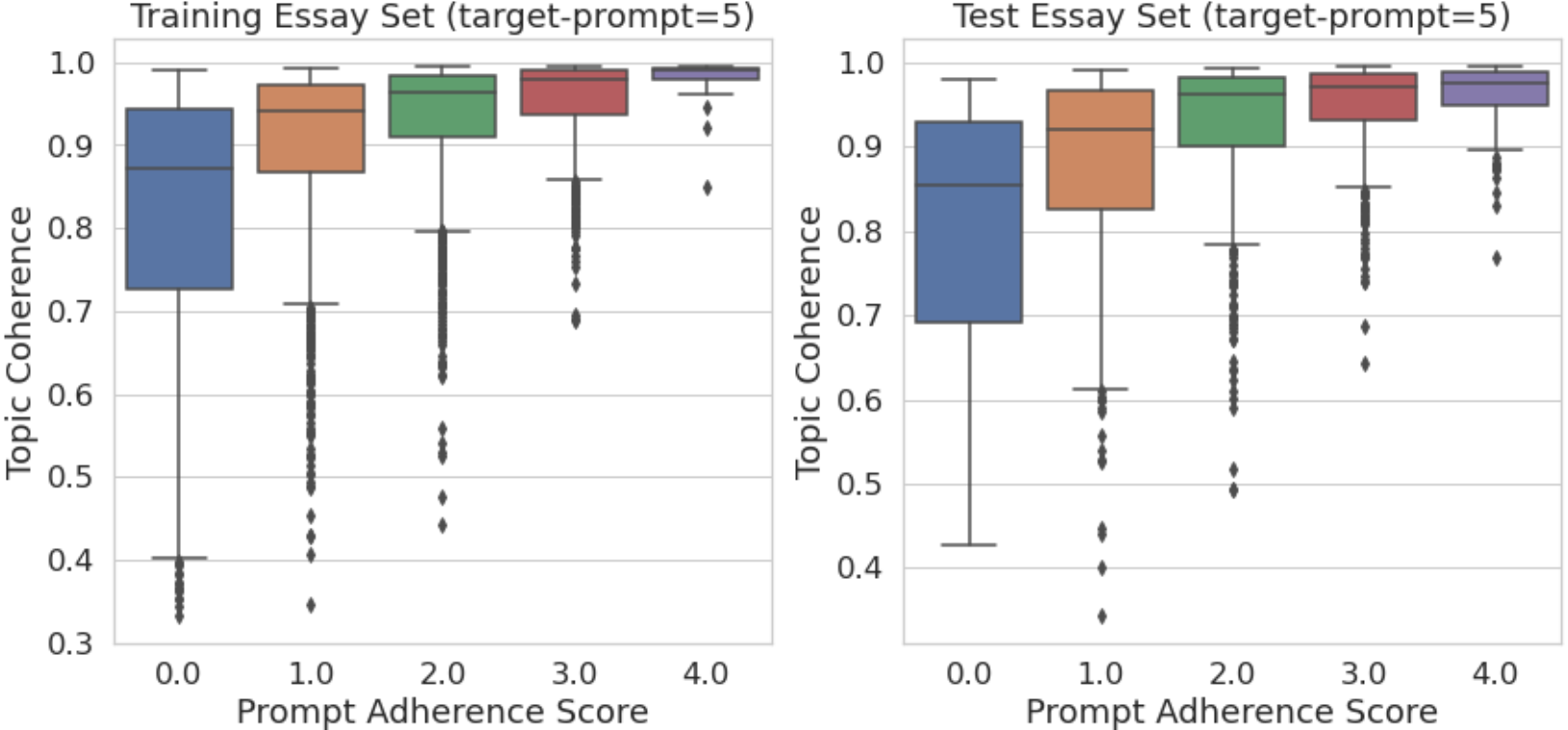}
    \caption{Box plot distributions of topic-coherence features of essays according to their \textit{Prompt Adherence} trait scores.}
    \label{fig7}
\end{figure}
In the main paper, we examined the relationship of our topic-coherence feature with \textit{Narrativity} trait score to see if extracted features using LDA truly reflect the coherence of the essay (Section~\ref{sec3.2}). Since we subsequently found that the topic is highly related to the prompt, we additionally investigated the feature relationship with the \textit{Prompt Adherence} trait score, which is another coherence-related trait \citep{shin2022evaluating}. We also examine the case of predicting target-prompt 5, where the training set contains essays of prompts except 5. Since only prompts 3--6 have \textit{Prompt Adherence} trait for evaluation, plotted training set only contains essays of prompts 3,4 and 6. Figure~\ref{fig7} shows similar tendencies as the distribution with \textit{Narrativity} trait, implying that the topic-coherence feature also conveys whether the essay written adherent to the prompt. These findings further explain the observed significant improvements on \textit{Prompt Adherence} trait scoring, in incremental analysis (Figure~\ref{fig5}). 


\section{Examples of the Prompt}
\label{sub:appendix2}

Table~\ref{tab10} shows the specific examples of the prompt in the ASAP dataset, which we utilized. We encoded the corresponding prompt contents for each essay. Prompts 1--2 define argumentative essay writing, prompts 3--6 describe the writing of source-dependent essays, and prompts 7--8 define the narrative type of essays. 

\begin{table}[h]
\renewcommand{\arraystretch}{1.4}
\centering
\begin{small}
\scalebox{
0.8}{
\begin{tabular}{c|p{7.5cm}}
\hline
Prompt ID & Prompt\\
\hline
1 & More and more people use computers, but not everyone agrees that this benefits society. Those who support advances in technology believe that computers have a positive effect on people. They teach hand-eye coordination, give people the ability to learn about faraway places and people, and even allow people to talk online with other people. Others have different ideas. Some experts are concerned that people are spending too much time on their computers and less time exercising, enjoying nature, and interacting with family and friends. Write a letter to your local newspaper in which you state your opinion on the effects computers have on people. Persuade the readers to agree with you.\\
\hline
2 & Censorship in the Libraries. "All of us can think of a book that we hope none of our children or any other children have taken off the shelf. But if I have the right to remove that book from the shelf -- that work I abhor -- then you also have exactly the same right and so does everyone else. And then we have no books left on the shelf for any of us." --Katherine Paterson, Author. Write a persuasive essay to a newspaper reflecting your vies on censorship in libraries. Do you believe that certain materials, such as books, music, movies, magazines, etc., should be removed from the shelves if they are found offensive? Support your position with convincing arguments from your own experience, observations, and/or reading.\\
\hline
3 & Write a response that explains how the features of the setting affect the cyclist. In your response, include examples from the essay that support your conclusion.\\
\hline
4 & Read the last paragraph of the story. "When they come back, Saeng vowed silently to herself, in the spring, when the snows melt and the geese return and this hibiscus is budding, then I will take that test again." Write a response that explains why the author concludes the story with this paragraph. In your response, include details and examples from the story that support your ideas. \\
\hline
5 & Describe the mood created by the author in the memoir. Support your answer with relevant and specific information from the memoir. \\
\hline
6 & Based on the excerpt, describe the obstacles the builders of the Empire State Building faced in attempting to allow dirigibles to dock there. Support your answer with relevant and specific information from the excerpt. \\
\hline
7 & Write about patience. Being patient means that you are understanding and tolerant. A patient person experience difficulties without complaining. Do only one of the following: write a story about a time when you were patient OR write a story about a time when someone you know was patient OR write a story in your own way about patience. \\
\hline
8 & We all understand the benefits of laughter. For example, someone once said, "Laughter is the shortest distance between two people." Many other people believe that laughter is an important part of any relationship. Tell a true story in which laughter was one element or part.\\
\hline
\end{tabular}}\end{small}
\caption{\label{tab10}
The eight prompts of the ASAP dataset.}
\end{table}

\end{document}